\documentclass{article}

    \usepackage[final]{neurips_2020}

\usepackage[utf8]{inputenc} 
\usepackage[T1]{fontenc}    
\usepackage{url}            
\usepackage{booktabs}       
\usepackage{amsfonts}       
\usepackage{nicefrac}       
\usepackage{microtype}      
\usepackage{graphicx}
\usepackage{amsmath}
\usepackage{booktabs}
\usepackage{multirow}

\usepackage{babel}
\usepackage{caption}
\usepackage{subcaption}

\title{Sentiment Analysis: Predicting Yelp Scores}

\author{%
  Bhanu Prakash Reddy Guda\quad Mashrin Srivastava\quad Deep Karkhanis\\
  Carnegie Mellon University\\
  \texttt{\{bguda,mashrins,dkarkhan\}@cs.cmu.edu} \\
}

\begin{document}

\maketitle

\begin{abstract}
  In this work, we predict the sentiment of restaurant reviews based on a subset of the Yelp Open Dataset. We utilize the meta features and text available in the dataset and evaluate several machine learning and state-of-the-art deep learning approaches for the prediction task. Through several qualitative experiments, we show the success of the deep models with attention mechanism in learning a balanced model for reviews across different restaurants. Finally, we propose a novel Multi-tasked joint BERT model that improves the overall classification performance. 
\end{abstract}

\section{Introduction}
Analyzing online reviews is crucial for many local and large-scale businesses that aim to excel by prioritizing customer satisfaction. Positive feedback from customers may prosper the store businesses, while negative one could have opposite consequences. Manually going through the reviews for understanding the overall tone could be a laborious process. An \textit{automated sentiment analysis} approach that examines the review and predicts its tone by looking at its structured (meta features) and unstructured information (review text) could benefit the businesses in quickly filtering and emphasizing their focus on improving the negative reviews. While several datasets have product reviews associated with them, in this work, we use the reviews submitted on the Yelp platform, a platform with an open dataset that has proven to be good and accurate for research purposes, to achieve this objective in a restaurant setting.

Among the several attributes that are available in the Yelp data, we focus mostly on the star rating, review text, some useful meta features, and sentiment score values. We organize the rest of the paper as follows. We provide a detailed description of the feature analysis and feature selection in Section \ref{sec:dataset} and build towards a novel sentiment prediction model in Section \ref{sec:methods} by progressively evaluating several machine learning baselines and the key ideas of prominent milestones in the field of deep learning approaches for natural language processing. We provide our intuitions behind using a particular architecture and consequently contribute a novel BERT based multi-tasked deep model with both (joint) meta features and review text as the input features for sentiment classification. In Section \ref{sec:results} we extensively evaluate the baselines along with the proposed model through quantitative and qualitative results.
\section{Related Work}
\label{sec:related}
Sentiment analysis is a popular and well-explored task in several domains. Among many others, the most common gold standard datasets that academia focuses on are the IMDb movie reviews~\cite{imdb}, Amazon product reviews~\cite{amazon}, Stanford Sentiment Treebank~\cite{sst}, and Yelp restaurant reviews~\cite{yelp-full-data}. Several of the proposed approaches in the past aim to learn a model specific to one dataset or a general model that performs well across all datasets; the latter one being more prevalent. The choice of model innovation to achieve competitive state-of-the-art performances among these works varies widely. \cite{wang2012baselines,socher2013recursive,thongtan2019sentiment} attempt to develop sophisticated ML models with careful feature engineering of meta data and review texts. \cite{kim,han,bert,xlnet} propose deep representations of the review text with an additional component of attending on parts of the input text to improve the overall performance. \cite{p1,p2} summarizes and reports the performances of these methods specifically on the Yelp dataset. 

Our objective of developing a model specifically for the Yelp review sentiment classification task enables us to differ from the prior approaches that learn a general language model and later fine-tune it for downstream tasks using only text. Our end-to-end novel architecture for the Yelp review sentiment classification task is inspired by the more recent idea of multi-tasked joint learning \cite{empath} for a closely related task of empathy prediction using text and meta attributes.

\section{Dataset and pre-processing}
\label{sec:dataset}
The dataset used for this paper is a subset of the Yelp Review dataset~\cite{yelp-full-data} which is a commonly used publicly available dataset for sentiment analysis. This dataset includes 8,021,122 reviews from 209,393 businesses in 10 metropolitan areas. This data is structured in JSON files, including business,
review, user, check-in, tip, and photo data. 
The subset of the data we have used has 36,692 customer reviews of restaurants. This data has a total of 42 features. However, some of these 36,692 reviews have some features missing. This dataset can be found at \cite{yelp-subset}.

Before the data could be used for training any machine learning model, it is necessary to do some pre-processing. Sentiment analysis is viewed as a supervised learning task. Of the 42 attributes provided in this dataset, two attributes: star rating and sentiment score are considered labels and the rest are taken as input features to define our supervised learning problem. \textbf{Star rating} -- Stars of the review. This is a typical one (worst) to five (best) scale which takes values as integers and is only available in the training data. \textbf{Sentiment score} -- A quantification of the text’s sentiment using the AFINN lexicon. The value ranges from -5 (very negative) to 5 (very positive).

\subsubsection*{Feature engineering} 

We split the data into train, validation and test set (70:15:15) for the purpose of training and evaluating our models. Moreover, we only use a subset of the features for training. Specifically, we \textbf{dropped majority of the binary features} indicating an occurrence of particular words like terrible, disgusting, amazing, \dots. This is because natural language models are prone to overfitting on such word occurrence features. Since natural language has a lot of freedom, most words can be used in most sentiments. Thus, models which do not rely on such bag-of-words approaches usually learn much better since they don't overfit on the training data's words. 

\begin{figure}[]
\centering
  \includegraphics[width=0.75\textwidth]{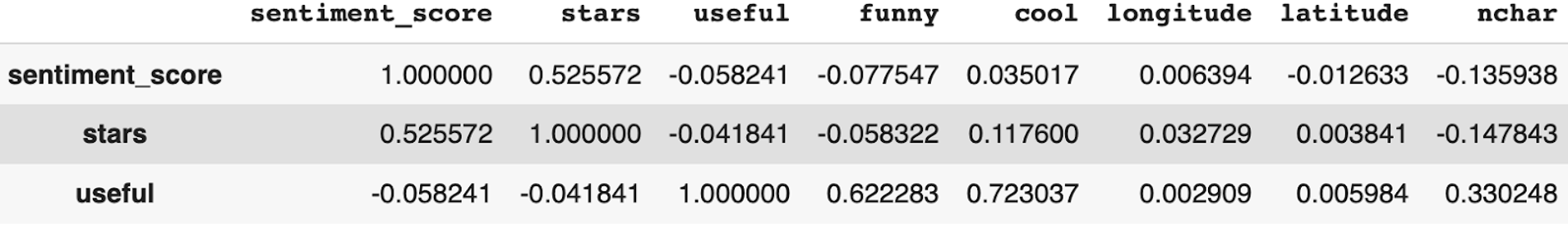}
  \caption{Correlation Matrix over features and labels}
  \label{fig:corr_mat}
\end{figure}

We also calculated the \textbf{correlation between all features} and labels in the train data. As seen in figure \ref{fig:corr_mat}, the ``stars'' feature or the start rating is highly positively correlated with the sentiment score. Hence, predicting the star rating gives us an idea of the sentiment of the review. Thus for the purpose of labeling positive vs negative sentiments, apart from the binary word occurrence indicators, we also \textbf{drop the sentiment score} and instead only use the star rating, hence we use the terms \textit{star rating prediction} and \textit{sentiment prediction} interchangeably. For this binary sentiment classification problem, our model labels a one to three star rating as a negative sentiment (0) and a four to five star rating as a positive sentiment (1). 

Further, we also drop features that clearly don't contain any useful sentiment information like reviewer name, number of characters and words in the review and the city name since most reviews are from the same city. Note that attributes that help identify the restaurant itself (like latitude and longitude) hurt the problem of sentiment analysis. This is because the model will simply start associating good restaurants to good reviews and not learn to infer sentiment from natural language.
We also convert the date to UNIX format, normalize the numerical attributes and convert the categories features to a one-hot encoding of 201 categories. For classical ML models, we also perform standard text pre-processing like removal of stop words, tokenization, converting to lower-case, lemmatization,  punctuation removal, and \textbf{TF-IDF calculations}~\cite{tfidf}. For deep learning models, we directly give the full text since we need to have the original sentence structure preserved.

\section{Methodology}
\label{sec:methods}
\begin{figure}  
    \centering
    \includegraphics[height=1.7cm]{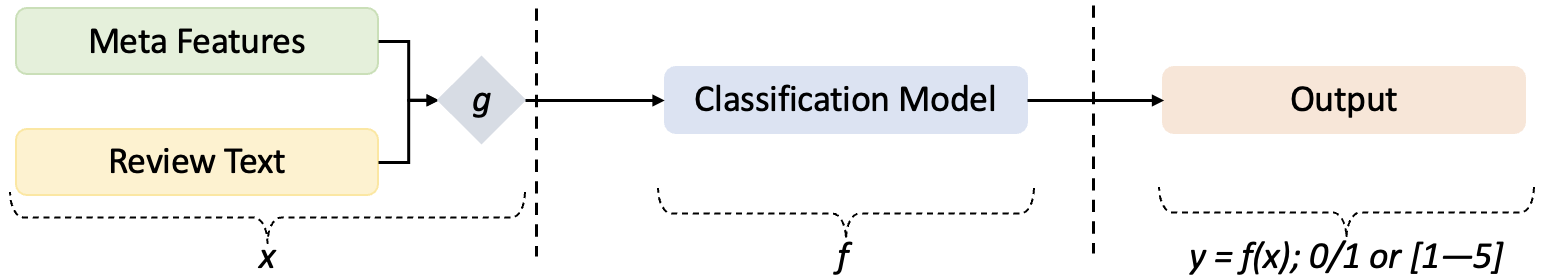}
    \caption{Classification setup used for the baselines and methods for our task. The function \textit{g} selects one of the feature sets or both of them by simply concatenating or applying a joint transformation.}
    \label{fig:setup}
\end{figure}

In this section, we describe various machine learning baselines and deep learning models used in our task. All our models follow the typical classification setup as shown in Figure \ref{fig:setup}, input features - model - output class. As mentioned in the dataset section, we have 2 sets of input features, \textbf{review text} and \textbf{meta features}. The output is either a binary classification i.e good star rating or a bad star rating, or a more granular 5-star classification i.e an integer in the range [1, 5]. For the model part, we provide our intuitions behind using a particular model, and in Section \ref{sec:results}, we validate our intuitions by analyzing the model interpretability.

\subsection{Machine Learning Baselines} 
In the first part of our experimentation, we apply 6 machine learning models, i.e Decision Trees (DT), Random Forest Classifier (RF), K-Nearest Neighbour Classifier (KNN), Support Vector Machines (SVM), Gradient Boosting Classifier (GB), and Multi-Layer Perceptron (MLP). To understand \textit{I1: Impact of the meta features and the review text individually and jointly on the classification task}, we experiment with 3 different setups of these models. (1) Using only the meta features as input for both binary as well as 5-star classification tasks. (2) Providing only the \textbf{review text} represented as a \textsc{tf-idf}~\cite{tfidf} vector as input. (3) Concatenating the \textsc{tf-idf} vector with meta features as input to each of these models.
In all the setups, we use the standard \textsc{scikit-learn}~\cite{sklearn} implementations for these models with hyperparameter tuning using the validation set.
\subsection{Deep Review Representation (DRR)}
While the simple machine learning models achieved good performance values (Table \ref{tab:results}) in both binary and 5-star classification setups, one of the bottlenecks is to represent convoluted information i.e review text through a simple \textsc{tf-idf} vector notation. To this end, we have experimented with a few notable advancements in deep learning techniques for better representation of text. Broadly the advancements have one of the following backbones as the underlying architecture: \textit{convolutional neural networks}, \textit{long short term memory networks}, and \textit{transformer} units. These networks are further boosted in their performance by a technique called \textit{attention}. Through these experiments, we try to understand \textit{I2: Impact of the deep high dimensional representations of the review text in the classification performance}. 

We will now describe the core components in each of these techniques.
\subsubsection{Convolutional Neural Networks}
In this experiment, we tackle the bottleneck by training convolutional neural networks (CNNs) trained on top of pre-trained word vectors. The word vectors are high-dimensional dense word representations that capture both global and local contexts. Our implementation of the CNNs is inspired by the model proposed by \cite{kim}. Given a review of length n (padded with zeros where necessary for facilitating matrix multiplications), a representation of the text is generated by applying convolution operations over various continuous windows using a \textit{filter} \textbf{w}. 

\begin{figure}
    \centering
    \includegraphics[height=5.5cm]{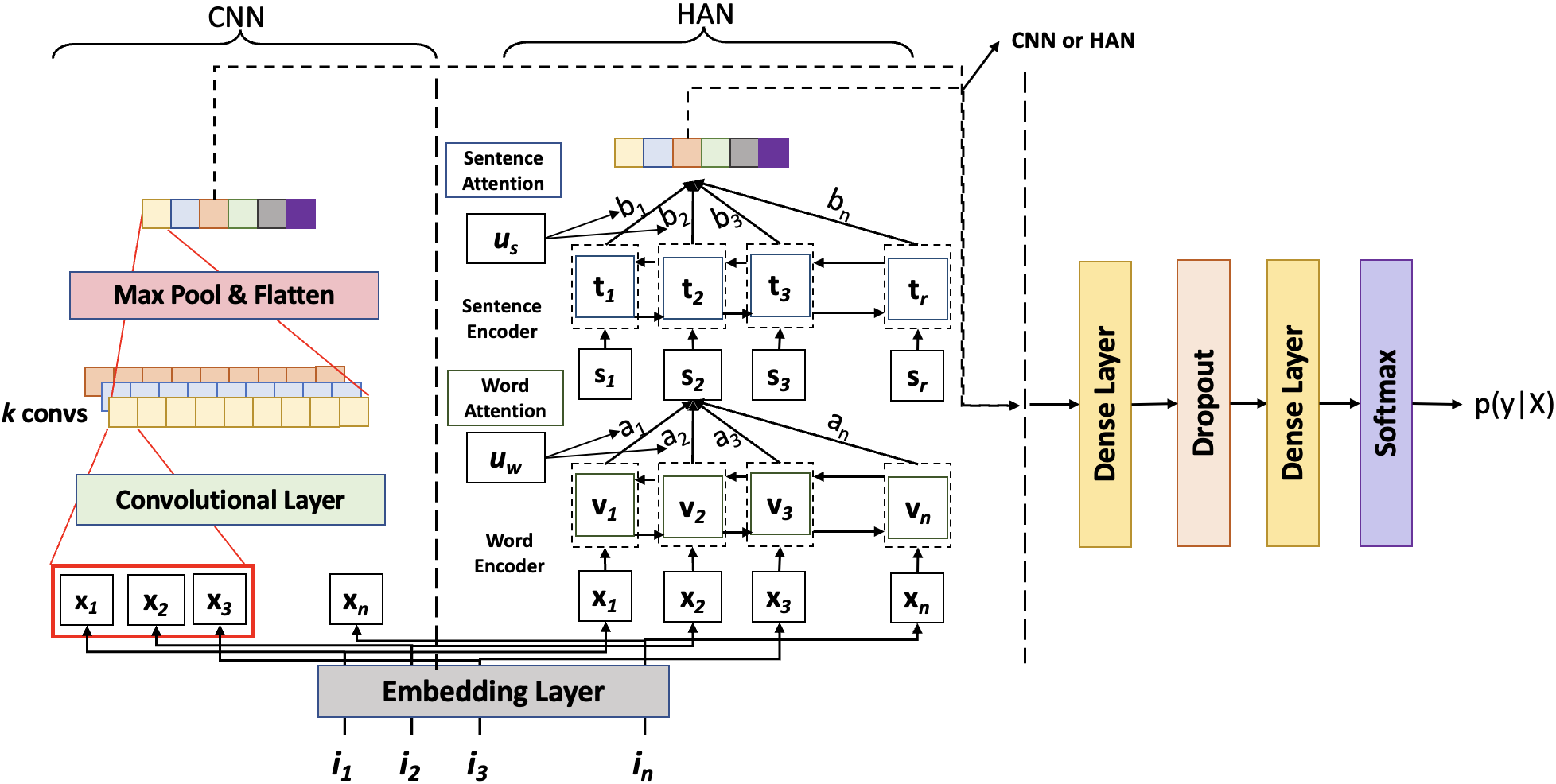}
    \caption{Left: Text encoding with CNN model; Middle: Text encoding with HAN model; Right: Final classification layers after textual representation.}
    \label{fig:cnn_lstm}
\end{figure}

In Figure \ref{fig:cnn_lstm}, $\mathbf{x}_i$ is the embedded representation of the $i^{th}$ word of sentence $s_2$, \textbf{k} convolution filters are applied, with a final max-over-time pooling operation \cite{collobert} to take max values per filter. The final concatenation of the max pool outputs is considered as the final \textit{DRR}. 
\subsubsection{Hierachical Attention Network (HAN)}
Our next experiment is more inspired by the structural nature of text ``words constitute sentences". \cite{han} proposed a hierarchical attention-based LSTM network for classifying text. The \textit{three components} that improve performance of the model over the plain CNN architecture are the \textit{LSTM} modules, \textit{hierarchical modeling}, and the \textit{attention} modules. As shown in Figure \ref{fig:cnn_lstm}-middle, one word is passed as input at each time step, and hidden state representation ($\mathbf{v_t}$) is computed for the next time step i.e. the next word in the LSTM unit.

The \textit{first component} - LSTMs~\cite{lstm} are chosen for their capability of remembering information over long sequences i.e. retaining required information about the $1^{st}$ word when dealing with the $n^{th}$ word. Since combination of words represent a sentence, the \textit{word encoder} is placed below the sentence encoder (\textit{second component}) as shown in Figure \ref{fig:cnn_lstm}-middle. The word encoder encodes the words $i_1\dots i_n$ to generate an overall representation of a sentence $s$ ($s_2$ in Figure \ref{fig:cnn_lstm}). Once all sentences obtain their representations from the same word encoder model, the sentence encoder generates an overall representation of the review. Sentence encoder is exactly similar to the word encoder, the only difference being sentences are the basic units rather than words. In both word and sentence encoders, we use the BiLSTM~\cite{bilstm} variant of LSTMs that capture information in both a left-to-right and right-to-left manner. 

The \textit{third component} that improved the performance of HAN is the attention component~\cite{attention}. A weight matrix ($\mathbf{u_w}$ for word encoder and $u_{s}$ for sentence encoder) is learnt that computes a score per unit representation, $\mathbf{a_l}$ for $l^{th}$ hidden time step representation $\mathbf{v_l}$ and $\mathbf{b_l}$ for $l^{th}$ hidden time step representation $\mathbf{t_l}$, and computes a weighted sum of hidden representations at each encoder level. The intuition is to weigh certain words and sentences more than others. We will cover this intuition in more detail in the BERT model section. Mathematically, the network can be represented as:
\begin{align}
    \mathbf{v}_{l} &= \overleftrightarrow{\text{LSTM}}(\mathbf{x}_l);\quad \mathrm{a}_l = \mathbf{u_w}.\mathbf{v}_l; \quad \mathbf{s}_k = \sum^{i=1}_{n} \mathrm{a}_i * \mathbf{x}_i \\
    \mathbf{t}_{l} &= \overleftrightarrow{\text{LSTM}}(\mathbf{s}_l);\quad \mathrm{b}_l = \mathbf{u_s}.\mathbf{t}_l; \quad DRR = \sum^{i=1}_{r} \mathrm{b}_i * \mathbf{t}_i
\end{align}
In Figure \ref{fig:cnn_lstm}-middle, $\mathbf{x}_1, \mathbf{x}_2,\dots\mathbf{x}_n$ are the embedded words of a sentence $\mathbf{s}_2$, $\mathbf{s}_1, \mathbf{s}_2,\dots\mathbf{s}_r$ are the encoded representations of the sentences of a review, $\overleftrightarrow{\text{LSTM}}$ is the Bi-directional LSTM module, and \textit{DRR} is the final deep representation of the review text.

\subsubsection{Deep Bidirectional Transformers - BERT}
\begin{figure}
    \centering
    \includegraphics[height=5.5cm]{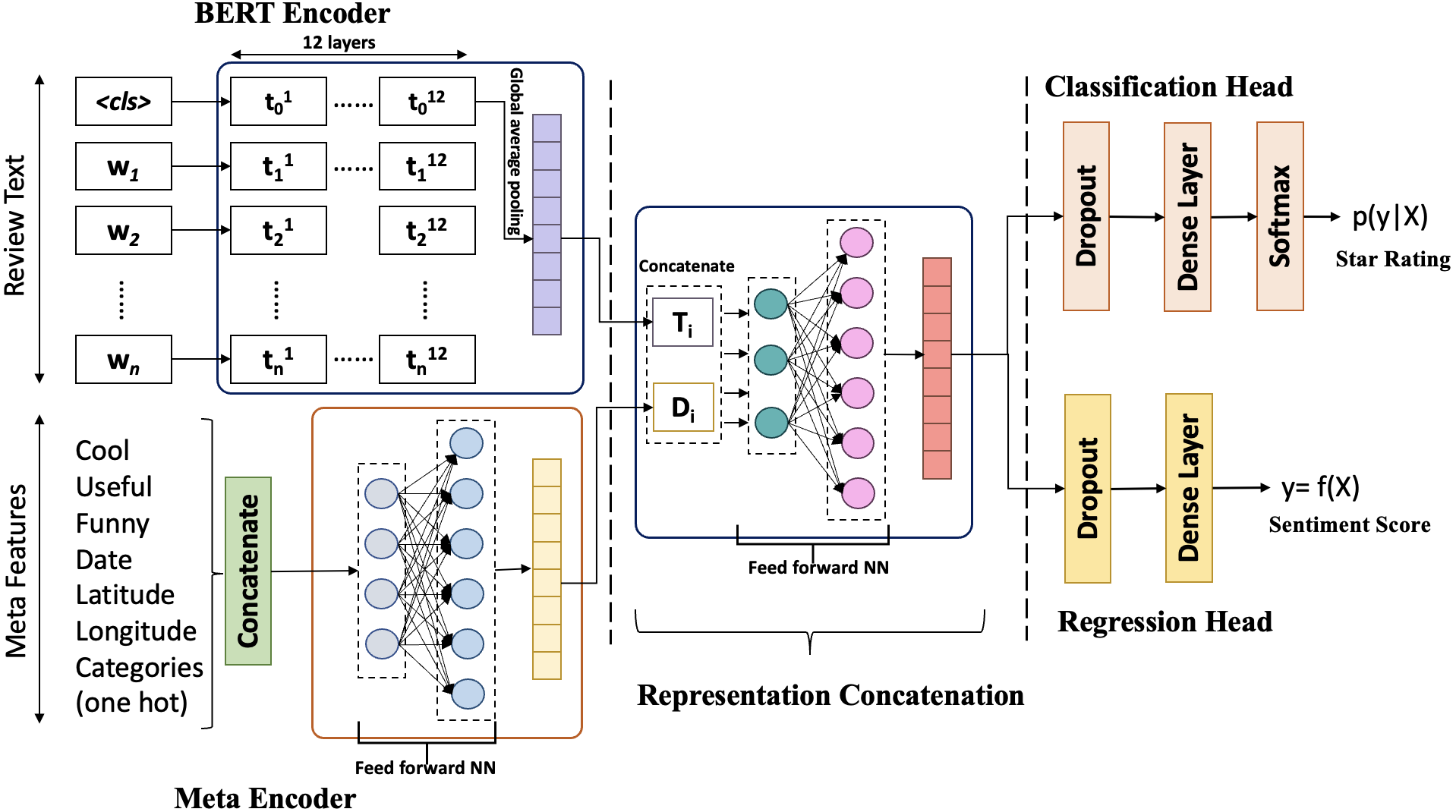}
    \caption{Top-Left: BERT encoder with bert-base-uncased backend; Bottom-Left: Encoding meta features through FNN; Middle: Concatenating text and meta representation and feeding to FNN; Top-right: Star rating classifier head; Bottom-right: Sentiment score regression head.}
    \label{fig:mt_dense}
\end{figure}
Another attempt we made in generating \textit{DRR} is the state-of-the-art language model BERT~\cite{bert}. Although newer variants of BERT like XLNET~\cite{xlnet}, RoBERTa~\cite{roberta},\dots achieve a better performance, we focused our efforts more on explainability of the model rather than quantitative improvements. We briefly explain the intuitions behind the architecture and its applicability in our context. We omit an exhaustive background description of the transformer units and the BERT architecture and refer readers to \cite{transformers}, \cite{bert} as well as a detailed guide \cite{jalammar}.

BERT is a language representation model, where deep bidirectional representation from unlabeled text is pre-trained on both left and the right context for masked language model and next sentence prediction tasks. Such a model, when pre-trained on large corpora of natural language texts can be used for wide tasks after being fine-tuned with just one additional output layer since the initial parameters of the model are optimized to learn general English language (language-specific BERT models are also available). The architecture of BERT allows for learning contextual word embeddings based on the \textit{context} by \textit{attending} to the other words in the input sentence. The attention mechanism enforces the model to assign a higher weightage to words that are relatively more important for the downstream task. 

In the standard implementation of BERT, the input is concatenated with a special token $<$\textit{cls}$>$ that captures the entire sentence representation after L self-attended transformer layers. We consider the Global Average Pooled output of the $<cls>$ representation as the dense representation of our review text. In the Figure \ref{fig:mt_dense}, the {\sc \textbf{BERT Encoder}} would generate a representation $\mathbf{T_i}$ for $i^{th}$ review text with words $\mathbf{w}_1,\mathbf{w}_2 \dots, \mathbf{w}_n$, which when fed to the {\sc \textbf{Classification Head}} would predict the probability of the review text belonging to a star rating class. Our intuition behind choosing attention-based models HAN and BERT among many competing deep learning architectures is because of our final intuition \textit{I3: Are there some signal words that indicate the sentiment of review and hence concentrating more on these terms would improve the classification performance?}
\subsection{Multi-Task Learning}
To reiterate, due to the strong positive correlation between the sentiment scores and the star rating, we refrained from using the sentiment score as an input to the machine learning models as it can cause data leak (final label) into the model. However, we could still utilize the sentiment scores to boost the performance of the aforementioned deep models through the concept of Multi-task learning(MTL)~\cite{mtl}. MTL states that joint learning of correlated tasks would improve the model by learning the shared representations in a joint space that would lead to a better generalization, in addition to boosting up the performance of one/both the tasks. 

Towards this end, we propose our \textit{first novel change} on top of the deep models by adding a classification head and regression head tasks for predicting the star rating class and estimating the sentiment score respectively. To facilitate this, in the Figure \ref{fig:cnn_lstm}, we add the {\sc \textbf{Classification}} and {\sc \textbf{Regression Heads}} from Figure \ref{fig:mt_dense} and pass the representation generating from CNN or HAN model to both the heads. For the joint loss optimization, we experiment with 2 variants. A simple variant is a weighted loss $\lambda*\mathcal{L}_1 + (1-\lambda)*\mathcal{L}_2$ where $\lambda$ is a hyperparameter. The second variant is the same as the weighted loss, with the exception that $\lambda$ is learned along with the model. We employ an uncertainty-based weighing of losses technique proposed in \cite{multiloss}. It is defined as 
\begin{align}
    \mathcal{L}_{joint} = \frac{1}{2\sigma_1^2}\mathcal{L}_1 + \frac{1}{2\sigma_2^2}\mathcal{L}_2 + \log(\sigma_1\sigma_2)
\end{align}
where $\sigma_1, \sigma_2$ are learnable parameters and $\mathcal{L}_1, \mathcal{L}_2$ are classification loss, cross-entropy, and regression losse, mean squared error, respectively. 
\subsection{Joint Multi-Task Learning}
After observing the contribution of both \textit{meta features} and the \textit{review text}, and the improvement when concatenated together in ML models, we experimented with concatenating the meta and text features in the dense models as well. Figure \ref{fig:mt_dense} represents our \textit{end-to-end novel architecture} with all the above mentioned modules. The overall pipeline is as follows: The text is encoded through the BERT encoder, the meta features are concatenated and passed through a feed-forward neural network (FNN) to generate the meta representation. The two representations, $\mathbf{T_i}, \mathbf{D_i}$, are then concatenated and passed through a FNN to learn the joint representation. Finally, the join representation is fed to the classification head (single task) or both heads (multi-task). Exact changes of Meta encoder and joint representation module are made to the LSTM and HAN models as well in Figure \ref{fig:cnn_lstm}. 
\subsection{Model Configurations}
We use the following hyperparameter configurations for the various ML baselines tested in this work: \textbf{Decision Trees} - \textit{gini entropy, no maximum depth}, \textbf{RF} - \textit{100 estimators, gini entropy, no max depth}, \textbf{SVM} - \textit{rbf kernel with C value 1.0}, \textbf{Gradient boosting} - \textit{learning rate 0.1, max depth of 5, max features of 0.5}, \textbf{KNN} - \textit{K=10}, \textbf{MLP} - \textit{2 hidden layers with 256, 128 units, learning rate 0.001, adam optimizer}. For deep models CNN and HAN, we use 100-dimensional GloVe embeddings~\cite{glove} for input words. In the CNN encoder, we use 256 1D filters of size 3 and stride 1. For the HAN model, we set the hidden size of word and sentence encoder LSTM units to 50, which implies that the input to the sentence encoder is of 50 dimensions. For the BERT encoder, we use the base uncased variant which has 12 hidden layers, 12 attention heads and hidden size 768 (embedding dim). 

For Meta encoder in Figure \ref{fig:mt_dense}, we use a single hidden layer FNN with 512 units followed by a dropout layer. For the concatenation module, we use a single layer FNN with 256 units. The dense layers in both classification and regression heads have the number of hidden units equal to the outputs i.e 2 (binary) or 5 (5-star) for star rating prediction and 1 for sentiment score estimation. Loss $\mathcal{L}_1$ is cross-entropy and loss $\mathcal{L}_2$ is mean squared error loss. In all the cases we use RELU as the  activation function and use a batch size of 32. The maximum sequence length is set to 50 for CNN and HAN, and 128 for BERT. We use Adam~\cite{adam} optimizer for CNN and HAN with learning rate 1e-4 and train for 5 epochs, and use AdamW~\cite{adamw} optimizer with learning rate 5e-5 and epsilon 1e-8 and fine-tune for 3 epochs in BERT variants.
\begin{figure}
    \centering
    \includegraphics[width=\textwidth]{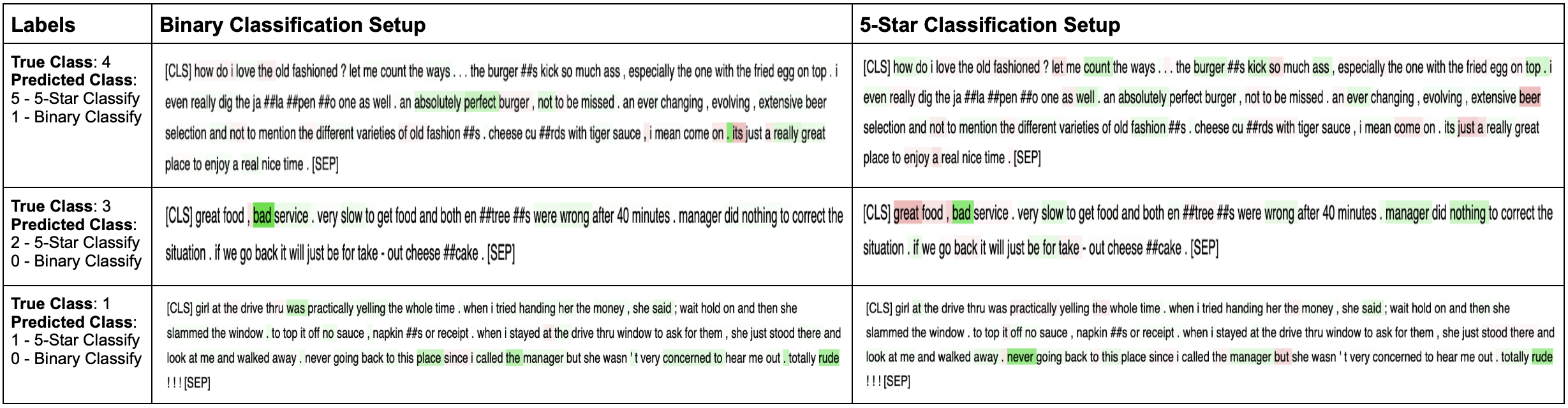}
    \caption{Qualitative examples with ground truth label and the BERT-MT-Joint model prediction in both binary and 5-star setup. Green color indicates that the word is weighted more in the correct class prediction (\textit{not positive class prediction}), and red indicate more weight in the opposite class prediction. Please zoom in for more clear visibility of the picture. }
    \vspace{-4mm}
    \label{fig:attention}
\end{figure}
\section{Results and Model Interpretability}
\label{sec:results}
In this section, we first provide the results of various models in different setups. In each of the model setups, we vary one or more of the input features, classification task, objective functions. While we report the results for both 5-star and binary star classification setups, the reader can assume the binary setup unless explicitly stated. Later, we provide an in-depth qualitative analysis of the models to evaluate the \textbf{intuitions} which we have specified earlier. We report the complete results of all setups in Table 1. In Table 1, Joint indicates meta features + review text as the input, 0/1 indicates binary mode and 5 indicates 5-star mode classification, and -MT indicates multi-tasked with sentiment score. The BERT-MT model outperforms all other models in all settings. However, multi-tasking when used together with Joint representation didn't improve the performance as expected, in fact in a couple of cases (CNN, BERT) it marginally decreased as well. We hypothesize that this is due to the increased zero or negative correlation that the sentiment score brings in when multi-tasked with star rating (Figure \ref{fig:corr_mat}). For all the follow-on experiments, we use the BERT-MT-Joint model or BERT-MT model if meta features are not available. 

\begin{minipage}{\textwidth}
  \begin{minipage}[b]{0.49\textwidth}
    \scalebox{0.63}{
    \begin{tabular}{c|c|c|c|c|c|c|c}
    \toprule
    & \multirow{3}{*}{Model}  & \multicolumn{6}{c}{Input Features}\\
    \cline{3-8}
    & & \multicolumn{2}{c|}{Meta Features} & \multicolumn{2}{c|}{Review Text} & \multicolumn{2}{c}{Joint}\\
    \cline{3-8}
    & & 0/1 & 5 & 0/1 & 5 & 0/1 & 5\\
    \midrule 
        \multirow{6}{*}{\rotatebox{90}{ML models}} & Decision Trees & 65.42 & 35.96 & 70.3 & 39.32 & 72.66 & 43.5\\
         & RF Classifier & 68.80 & 40.21 & 74.1 & 48.51 & 75.17 & 53.47\\
         & KNN & 63.74 & 34.2 & 69.5 & 38.5 & 70.92 & 46.42\\
         & SVM & \textbf{71.77} & \textbf{43.25} & 75.72 & 53.56 & 77.46 & 55.10\\
         & Gradient Boosting & 70.05 & 42.34 & 73.19 & 51.56 & 75.50 & 53.33\\
         & MLP & 69.99 & 41.99 & 73.22 & 51.25 & 75.85 & 52.01\\
    \midrule
    \multirow{6}{*}{\rotatebox{90}{Dense models}} & 
        CNN & -- & -- & 81.17 & 56.01 & 82.25 & 56.31\\
         & CNN-MT & -- & -- & 82.32 & 56.99 & 82.40 & 56.80\\
         & HAN & -- & -- & 84.59 & 58.33 & 85.98 & 59.25\\
         & HAN-MT & -- & -- & 85.91 & 59.14 & 85.95 & 59.21\\
         & BERT & -- & -- & 86.44 & 59.96 & 87.46 & 60.87\\
         & BERT-MT & -- & -- & \textbf{87.81} & \textbf{61.22} & \textbf{87.82} & \textbf{61.20}\\
    \bottomrule
    \end{tabular}
    }
    \label{tab:results}
    \captionof{table}{Test accuracy values in various setups}
  \end{minipage}
  \hfill
  \begin{minipage}[b]{0.49\textwidth}
    \centering
    \centering
        \includegraphics[scale=0.45]{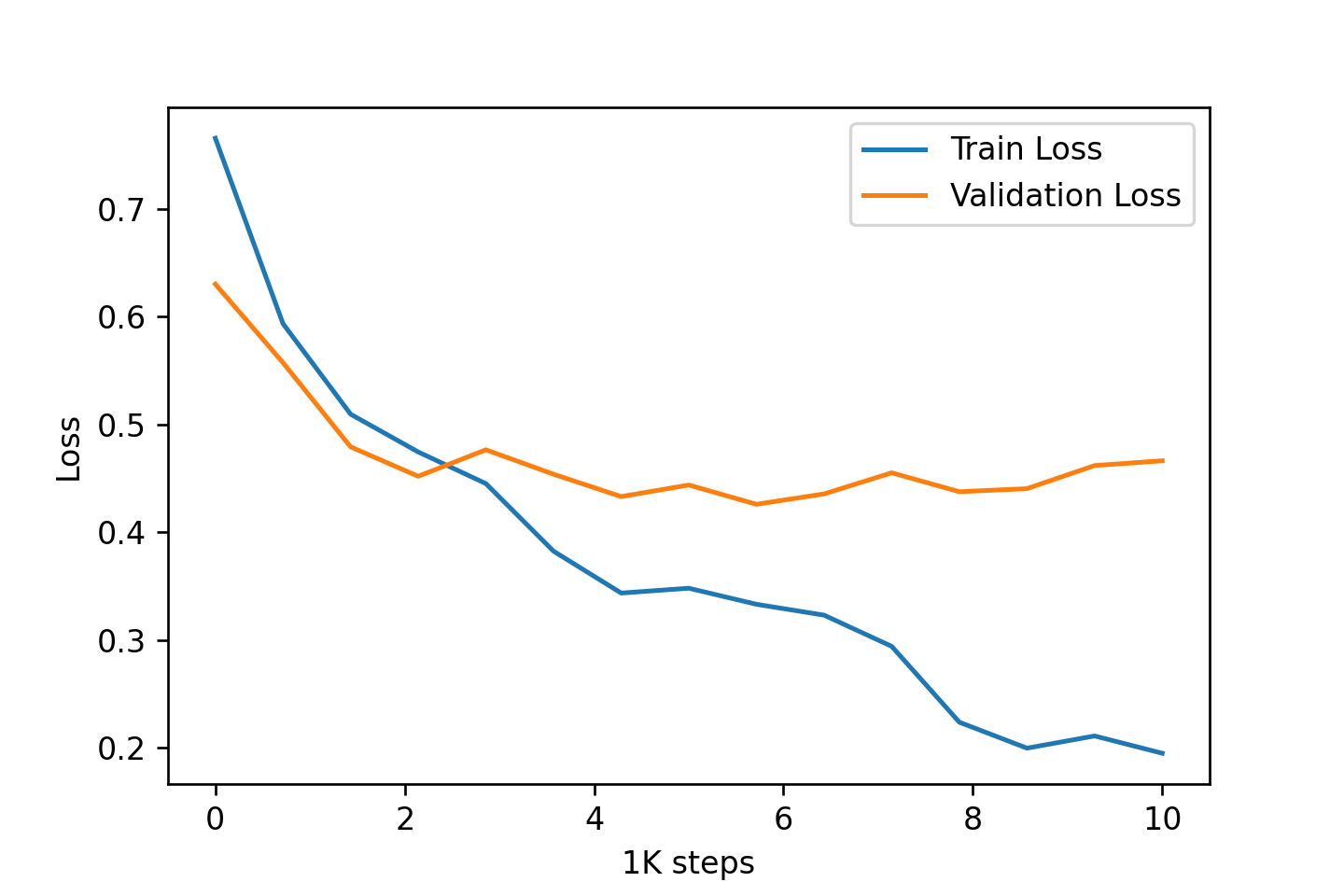}
    \captionof{figure}{Loss curves}
    \end{minipage}
\end{minipage}

\textit{\textbf{I1}: Impact of the meta features and the review text individually and jointly on the classification task} -- 
From the table, we can observe that both meta and text features are important in both 5-star and binary classification setups, with text features resulting in a better performance. When fed jointly, all the models report a better performance when compared to using either of the features alone. For deep models, since removing text reduces the model to a simple FNN, we report results for with text and with text + meta features. 

\textit{\textbf{I2}: Impact of the deep high dimensional representations of the review text in the classification performance} -- 
When compared against the simple tf-idf representation, the dense models achieve significantly better performance across all setups. Among the competing deep learning models, the contextual deep representation by the BERT model wins. 
\begin{figure*}[]
    \centering
    \begin{subfigure}[b]{\textwidth}
        \centering
        \includegraphics[scale=0.5]{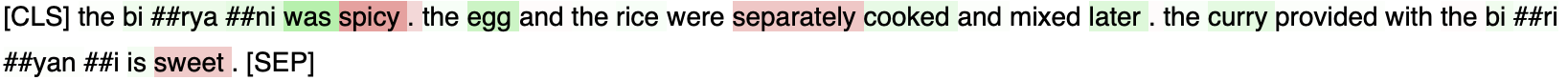}
        \caption{}
        \label{fig:rare_case}
    \end{subfigure}%
    \\
    \centering
    \begin{subfigure}[b]{0.49\textwidth}
        \centering
        \includegraphics[scale=0.33]{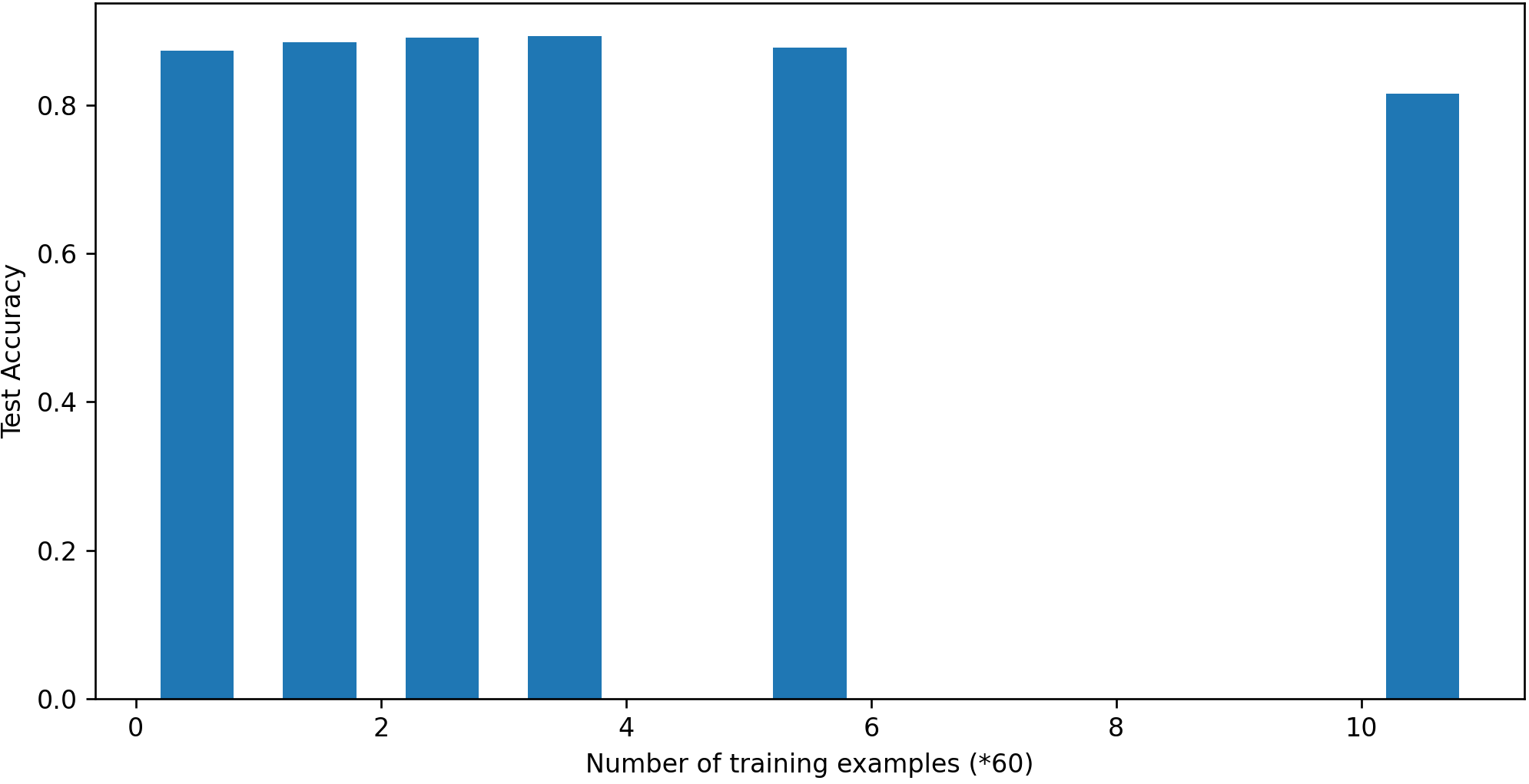}
        \caption{}
        \label{fig:bias}
    \end{subfigure}
    ~
    \begin{subfigure}[b]{0.49\textwidth}
        \centering
        \includegraphics[scale=0.25]{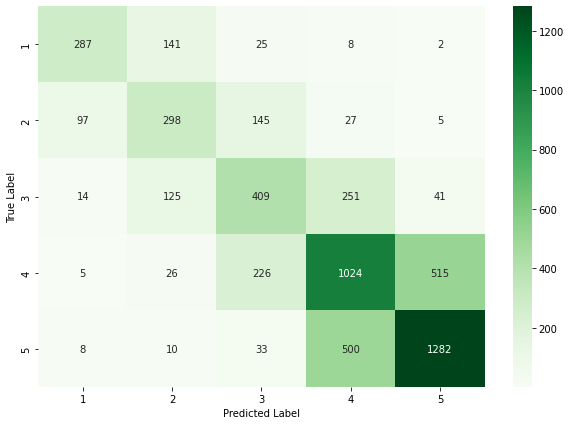}
        \caption{}
        \label{fig:cm}
    \end{subfigure}
    \caption{(a)An example where the BERT-MT is not able to classify correctly. (b) Test accuracy values of BERT-MT-Joint model on restaurants with different number of datapoints in the training set, (c) Confusion matrix in 5-star classification setup. F1 values are implied from the confusion matrix and hence we don't repeat in Table 1.}
    \vspace{-6mm}
\end{figure*}

\textit{\textbf{I3}: Are there some signal words that indicate the sentiment of the review and hence concentrating more on these terms would improve the classification performance?} -- Since there is no labeled dataset to generate quantitative results, we qualitatively analyzed the attention weights using the inverted gradients technique. Briefly, the IG method computes the input feature value multiplied by its gradient w.r.t loss i.e contribution of this word towards to either class prediction probability, for more details please refer to \cite{axiomatic}. In Figure \ref{fig:attention}, we report results for randomly chosen examples with different star ratings. As observed in the $2^{nd}$ column, for classifying the 4-star example, the model concentrated more on words like ``absolute, perfect, really, great" and words like ``bad,  rude" in 3-star and 1-star examples. Note that in binary classification, we attribute the 3-star as class 0 i.e negative sentiment. This experiment proves that there are certain words that act as signals for the classification task and attention-based models are indeed focusing more on the signal terms to achieve better performance.

\textit{\textbf{Q1}: 5-star vs binary classification} -- In this experiment, we wanted to observe the differences that are likely causing the poor performance of the 5-star classification. One can observe in Figure \ref{fig:attention} that for binary classification, the model simply concentrated on positive terms like ``absolutely perfect" to predict the class as positive. However, in a more nuanced setup, the model is concentrating on additional terms like ``kick ass, extensive, \dots" and therefore ended up classifying as 5-star. This difference is more pronounced in the 3-star example ($2^{nd}$ row of Figure \ref{fig:attention}). For simple negative classification (binary mode), the model was sufficient with the word ``bad service". However, in a 5-star setup, the attention is distributed between ``great food" and ``bad service, nothing", which indeed it is supposed to do, to find the neutrality. Due to assigning more weights to ``bad" and ``nothing", the model ended up in classifying the example as 2-star. For 1-star example, the sentiment is clear from the terms ``rude" and ``never" and hence in both binary and 5-star, the model was correct. This experiment proves that although the model is attending properly on the terms required, it is confusing between nearby classes and hence the reduced performance of a 5-star setup. We explore this confusion in more detail shortly.

\textit{\textbf{Q2}: Model Failure cases} -- While we report the superior performance of the BERT-MT-Joint model both quantitatively (Table 1) and qualitatively (Figure \ref{fig:attention}), to find failure cases, we manually explored by looking at the predictions of the model on a number of Google maps restaurant reviews. Out of the many good performance cases (binary setup), we found one type of example where our model is dominantly failing. In Figure \ref{fig:rare_case} we report one such example. Here the user specified a negative intention through the \textit{method of cooking} rather than using the standard signal words. In these extreme scenarios where an external knowledge is required, probably a good recipe for restaurants dishes, our model is significantly under-performing. 

\textit{\textbf{Q3}: Training convergence, bias, and label confusion} -- The last quantitative analysis that we perform pertains to observing the training convergence of the model, identifying any training biases, and labeling confusion in the model predictions. Figure 4 clearly shows that our training loss is decreasing with iterations and validation loss is saturated. Although we observe a slight increase in validation loss at the end, we employ an early stopping mechanism to avoid this. In Figure \ref{fig:bias} we plot the mean accuracy of performance on restaurants (identified by the \textit{name} attribute) with different number of examples in training. We observe a balanced performance across restaurants that have zero or low number of examples in training vs restaurants that have a significant presence in the training set. Finally in Figure \ref{fig:cm} we plot the confusion matrix in the nuanced 5-star classification setup. We observe that the model is mostly confusing between nearby classes, 1 \& 2, 2 \& 3, 3 \& 4, and 4 \& 5. We already discussed an intuition behind this confusion earlier in \textit{\textbf{Q1}}.

\section{Conclusion and Future Work}
From our extensive evaluations, we conclude that (1) jointly using meta and review text is always beneficial, (2) deep representation of text significantly boosts performance, (3) multi-tasking with sentiment score gives additional improvements, (4) attention mechanisms can enhance performance and interpretability (5) current training setup induces no bias in spite of reviews being from a small number of restaurants. The model is well-generalized with balanced performance.

We also point some future directions of exploration: (1) understanding the nuances between the nearby classes and consequently to devise a better performing model for 5-star classification, (2) incorporating external knowledge using knowledge graphs or commonsense reasoning, and (3) generalizing the model beyond restaurant reviews. 
\bibliographystyle{plainnat} 
\bibliography{main} 






\end{document}